# Similarità per la ricerca del dominio di una frase

*Metodo di valutazione del dominio di appartenenza di una frase*
Massimiliano Morrelli, Giacomo Pansini, Massimiliano Polito
Network Contacts - Via Olivetti 17 Molfetta (Ba)
Arturo Vitale
React Consulting - Via Alessandro Severo 58 Roma (Ro)

## Abstract

**English**. This document aims to study the best algorithms to verify the belonging of a specific document to a related domain by comparing different methods for calculating the distance between two vectors. This study has been made possible with the help of the structures made available by the Apache Spark framework. Starting from the study illustrated in the publication "New frontier of textual classification: Big data and distributed calculus" by Massimiliano Morrelli et al., We wanted to carry out a study on the possible implementation of a solution capable of calculating the Similarity of a sentence using the distributed environment.

**Italiano.** Il presente documento persegue l'obiettivo di studiare gli algoritmi migliori per verificare l'appartenenza di un determinato documento a un relativo dominio tramite un confronto di diversi metodi per il calcolo della distanza fra due vettori. Tale studio è stato condotto con l'ausilio delle strutture messe a disposizione dal framework Apache Spark. Partendo dallo studio illustrato nella pubblicazione *"Nuova frontiera della classificazione testuale: Big data e calcolo distribuito"* di Massimiliano Morrelli et al., si è voluto realizzare uno studio sulla possibile implementazione di una soluzione in grado di calcolare la Similarità di una frase sfruttando l'ambiente distribuito.

# 1. Stato dell'arte

La Statistica Testuale è una procedura che applica uno strumento quantitativo quale l'approccio statistico allo studio qualitativo di una qualunque tipologia di testo scritto. Si individuano due fasi distinte nel processo di analisi di un testo: l'analisi semantica e l'analisi di contenuto.

## 1.1 Analisi Semantica

L'analisi semantica, insieme all'analisi statistica delle parole e dei cluster, si propone di analizzare la struttura di uno scritto evidenziando le relazioni complesse che esistono tra le parole, mettendo in risalto la ricchezza del vocabolario, la distribuzione delle parole in un testo e la presenza di alcune strutture (ad esempio: costrutti sintattici, sinonimi, parole composte e segmenti ripetuti).

## 1.2 Analisi del Contenuto

Successivamente all'analisi semantica, fase preliminare, si effettua l'analisi di contenuto. Questa consiste nel classificare in temi i differenti segmenti di testo; un segmento di testo è un gruppo di parole che possiede da sé un significato. Raggruppando poi i temi in sotto-temi e mega-temi è possibile comprendere il senso, il significato e il messaggio evincibile dal testo, che spesso non è immediato, soprattutto per quanto riguarda le risposte a questionari aperti.
Con l'analisi di contenuto, è possibile classificare le idee entro un albero gerarchico.

## 1.3 Procedura Statistiche

Le procedure statistiche che stanno alla base dell'analisi semantica e di contenuto sono essenzialmente tre:

### 1.3.1 Costruzione del vocabolario

Il primo passo consiste nell'individuare le unità lessico-metriche che a seconda dei casi possono consistere in parole, parole composte, insiemi di parole o addirittura frasi; quest'insieme costituisce il vocabolario.

### 1.3.2 Lemmatizzazione

Quest'operazione consente di snellire il vocabolario di un testo tramite la riunificazione di forme lessicali corrispondenti a diverse flessioni di uno stesso lemma. In questo modo, per esempio, ogni forma verbale viene ricondotta all'infinito ed ogni sostantivo al maschile singolare. I lemmi costituiscono le occorrenze cioè le entità che ricorrono nei testi, rispettivamente le prime costituite da una parola, le seconde da un numero superiore di termini.

### 1.3.3 Numerotizzazione

Tramite un supporto informatico, è possibile effettuare la numerotizzazione, che consiste nell'attribuire una o più cifre ad ogni occorrenza ricavata nella fase di lemmatizzazione, contraddistinguendo con un certo contatore numerico la corrispondente unità alfanumerica. Esistono vari metodi di numerotizzazione: ad esempio è possibile associare ad ogni costrutto una serie di 0 e di 1 che indicano la presenza o l'assenza di una certa forma lessicale. Mettendo poi tutti i numeri in una stringa è possibile effettuare diverse elaborazioni statistiche.

## 1.4 Elaborazioni Statistiche

Dopo la numerotizzazione è possibile elaborare i dati in vario modo, di seguito vengono elencate diverse metodologie che vengono applicate nello studio dei documenti.

Nel nostro studio abbiamo sfruttato i concetti descritti nel sotto paragrafo *"Matrice delle distanze"*, ma non è il metodo che è stato utilizzato.

### 1.4.1 Tabella delle frequenze

Dal numero di volte che un'occorrenza si ripete nello scritto analizzato, è possibile creare semplici tabelle che evidenziano i concetti dominanti presenti nel testo. Allo stesso tempo è utile poiché permette di eliminare le forme con scarsa frequenza e di visualizzare quelle invece più ricorrenti, cioè le parole-chiave.

### 1.4.2 Tabella delle contingenze

E' una matrice di tipo booleano le cui righe sono costituite da raggruppamenti (cluster) delle parole-chiave, mentre le colonne sono costituite dalle lemmatizzazioni effettuate in precedenza.
La tabella è riempita di " 1 " o " 0 ", rispettivamente se l'occorrenza esaminata appartiene o no al cluster corrispondente. Questa tabella è utile per costruire una serie di sequenze binarie che connotano quella parte del campione esaminato (allo stesso modo, se consideriamo la dimensione verticale, si rappresentano identikit delle lemmatizzazioni).

### 1.4.3 Matrice delle distanze

Considerando le righe a due a due delle tabelle delle contingenze, è possibile definire un indice di similarità, cioè un numero che quantifichi la distanza tra le stesse. Si costruisce una matrice quadrata dove gli elementi numerici all'interno indicano quanto due sottogruppi sono simili fra loro.

### 1.4.4 Clustering gerarchico ascendente

Basandosi sulle matrici delle distanze è possibile aggregare, a due a due, le varie categorie del campione esaminato o le lemmatizzazioni precedentemente ottenute. I due sottogruppi più vicini andranno a costituire una nuova identità unica, la cui distanza dagli altri è pari alla media delle distanze che gli stessi avevano prima della fusione. Così facendo, ad ogni passo, il totale degli oggetti di partenza si riduce di un'unità.

### 1.4.5 Analisi delle corrispondenze

Dalla matrice delle distanze, sovrapponendo la mappatura dello spazio delle categorie costituenti il campione esaminato sulla mappatura delle lemmatizzazioni di partenza, è possibile visualizzare le corrispondenze tra le rispettive collocazioni e dunque un legame reciproco. Più le occorrenze appaiono spesso nelle risposte più è grande la vicinanza nel grafico tra queste e le relative categorie costituenti il campione. Quest'analisi è molto efficace nel trattamento delle risposte ai questionari aperti.

## 2. Calcolo della similarità

### 2.1 Similarità

Sia $E$ l'insieme di entità con caratteristiche comuni, è possibile definire la funzione che ad ogni coppia di elementi $e_1, e_2 \in E$ associa un numero reale compreso tra 0 e 1. Tale valore è noto con il nome di *Coefficiente di similarità*.

$$sim: E \times E \rightarrow [0,1]$$

L'algoritmo scelto per valutare la similarità di un documento e quindi calcolare il coefficiente di similarità è il K-nearest neighbors (KNN), descritto nel paragrafo successivo.

### 2.2 K-nearest neighbors

L'algoritmo *k-nearest neighbors* (KNN) è un algoritmo di apprendimento automatico supervisionato utilizzato per risolvere sia problemi di classificazione che di regressione.

Un algoritmo di apprendimento automatico supervisionato si basa su dati di input etichettati per addestrare una funzione che produce un output appropriato quando vengono forniti nuovi dati senza etichetta, ovvero, data una coppia di valori $(x, y)$ sarà necessario trovare una relazione fra di loro. In particolare si dovrà addestrare una funzione $h = X \rightarrow Y$ che permetterà di poter conoscere il valore y tramite la funzione $h(x)$.

L'algoritmo KNN presuppone che oggetti simili esistano nelle immediate vicinanze, in altre parole, cose simili sono vicine l'una all'altra. L'algoritmo KNN dipende da questo presupposto, e riesce ad unire il concetto di somiglianza (a volte chiamata distanza, prossimità o vicinanza) e reinterpretarlo in una funzione matematica.

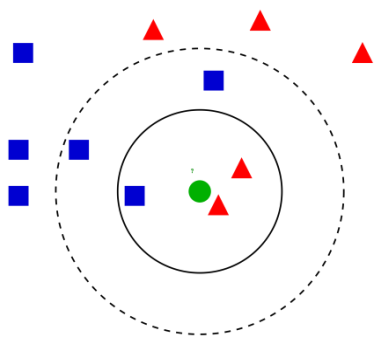

*Figura 2.1 - k-esima suddivisione per la classificazione k-nn di uno spazio contenente elementi*

Il classificatore KNN è anche un algoritmo di apprendimento non parametrico e basato su istanze. Il significato di non parametrico sta nella eliminazione delle assunzioni esplicite sulla formula funzionale di h, così facendo si evita il pericolo di modificare la distribuzione sottostante dei dati.

L'apprendimento basato su istanze, implica che l'apprendimento non avviene tramite l'addestramento esplicito di un modello ma si sceglie di calcolare a runtime in adeguate strutture dati, le istanze[1] di addestramento che costituiscono la "conoscenza" utile alla predizione (1).

---

[1] Istanza: punto in uno spazio n-dimensionale.

L'algoritmo KNN si riduce essenzialmente a valutare la somiglianza degli elementi più vicini e, a maggioranza, classificare l'elemento. La somiglianza viene definita in base a una metrica di distanza tra due punti di dati. Una scelta popolare è la distanza euclidea data da:

$$d(x, x') = \sqrt{(x_1 - x'_1)^2 + (x_2 - x'_2)^2 + \cdots + (x_n - x'_n)^2} \qquad Eq.2.1$$

ma è possibile valutare misure più adatte al proprio dataset ed è possibile scegliere tra:

- Manhattan;
- Chebyshev;
- Hamming.

Nel paragrafo successivo verranno analizzati i risultati ottenuti dagli algoritmi di misura messi a disposizione dal KNN.

## 2.3 Metodi per il calcolo della distanza - La Similarità del coseno

La similarità del coseno, o cosine similarity, è una tecnica euristica per la misurazione della similitudine tra due vettori effettuata calcolando il coseno tra di loro, usata generalmente per il confronto di testi nel data mining e nell'analisi del testo.

La somiglianza del coseno tra due vettori (o due documenti nello Spazio vettoriale) è una misura che calcola il coseno dell'angolo tra loro. Questa metrica è una misura dell'orientamento e non della grandezza, può essere vista come un confronto su uno spazio normalizzato perché non stiamo prendendo in considerazione solo la grandezza di ogni conteggio delle parole di ciascun documento, ma l'angolo tra i documenti. Quello che dobbiamo fare per costruire l'equazione della somiglianza del coseno è risolvere l'equazione del prodotto punto per $cos\theta$:

$$\vec{a} \cdot \vec{b} = \|a\|\|b\| \cos\theta \qquad Eq.2.2$$

$$\cos\theta = \frac{\vec{a} \cdot \vec{b}}{\|a\|\|b\|} \qquad Eq.2.3$$

Ogni documento verrà rappresentato come tutti gli attributi (parole) che può avere e registreremo la frequenza con cui ogni attributo si presenta nel singolo documento. Possiamo così rappresentare ogni documento come un vettore. Nel nostro caso la similarità del coseno potrebbe essere così rappresentato

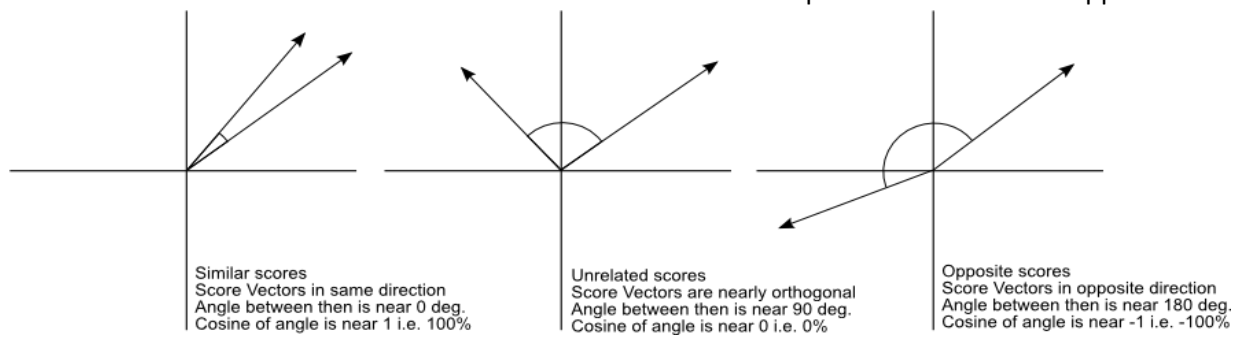

*Figura 2.2 - Calcolo della distanza del coseno*

Avremo quindi documenti simili caratterizzati da un angolo prossimo a zero ed un coseno prossimo a 1.

Stante la rappresentazione di ciascun documento tramite la frequenza (occorrenza) avremo una rappresentazione in forma vettoriale costituita da vettori aventi solamente numeri positivi e dunque, in una rappresentazione grafica questi vettori saranno sempre nel primo quadrante di un piano. il coseno non sarà mai negativo visto che il massimo angolo che potremo osservare non sarà mai superiore a 90°.

Questa misura è particolarmente appropriata per i nostri scopi poiché non tiene conto della lunghezza del vettore ma del suo orientamento e questo coincide con il nostro scenario

La scelta della distanza, utilizzando la similarità del coseno, ricade non solo su una valutazione delle performance comparata con gli altri metodi, che come si evince dalla Tabella 2.1, la similarità del coseno restituisce una accuracy migliore.

| Distanza Metodo | Euclidea | Manhattan | Canberra | Coseno |
|---|---|---|---|---|
| 1-NN | 97.32% | 97.28% | 98.07% | 98.40% |
| 2-NN | 94.08% | 94.32% | 96.29% | 96.62% |
| 3-NN | 91.31% | 91.50% | 95.96% | 96.24% |

*Tabella 2.1 – Calcolo delle performance delle varie distanze di KNN*

Notiamo anche come gli altri metodi valutino sempre la magnitudine dei vettori e non l'orientamento, risultando quindi meno adatti nel cogliere il senso generale di una frase. Infatti qualora comparassimo due frasi tra di loro simili come concetto, potremmo avere casi in cui esse risultino molto simili sotto il punto di vista dell'angolazione (distanza del coseno) ma molto distanti secondo la distanza euclidea (o altre distanze basate sulla magnitudine).

Prendiamo ad esempio il caso in cui due frasi che abbiano il medesimo senso siano di diversa lunghezza ed una presenti una parola chiave ripetuta molte più volte rispetto all'altra. In questo caso le due frasi risulterebbero molto distanti secondo la distanza euclidea ma molto vicini secondo la distanza del coseno.

## 2.4   Logiche e implementazione

In ogni lingua ci sono molteplici modi per esprimere un concetto. Normalmente i termini sono portatori di forte ambiguità ed è necessario interpretare correttamente l'intera frase per comprenderne e per carpirne il senso.

Un passo preliminare sarà quello di eliminare tutte quelle parole, che non aggiungono niente al significato intrinseco della frase ma sono molto frequenti nei documenti. Questo può creare distorsioni negli algoritmi. Di conseguenza sarà necessario identificare le cosiddette stop-words, queste "parole d'ordine" potrebbero non aggiungere valore al significato del documento.

Con la rimozione delle stop-words si ha di conseguenza la riduzione delle dimensioni del set di dati e quindi si riduce anche il tempo di addestramento del modello; inoltre le prestazioni otterranno un miglioramento, in quanto essendo rimasti meno token significativi, la precisione della classificazione risulterà più precisa; avremo così una trasformazione di una generica frase nelle sue componenti principali che rappresentano solo il concetto.

Gli algoritmi di NLP – Natural Language Processing - sono in grado di elaborare solo dati numerici e non stringhe testuali. Di conseguenza risulta necessario pre-processare i documenti tramite un modello che riceva in input i dati di tipo testuale (le frasi) e restituisca il relativo vettore delle occorrenze.

Pertanto, il modello Bag of Words viene utilizzato per pre-processare il testo, convertendolo in un bag of words, che tiene conto del numero totale di occorrenze delle parole utilizzate più di frequente.

## 2.5   Modello Bag of words

Bag of Words (BoW) è un algoritmo che conta quante volte una parola appare in un documento. Il conteggio delle parole consente di confrontare i documenti e valutare le loro somiglianze per applicazioni come la ricerca, la classificazione dei documenti e la modellazione degli argomenti. BoW è anche un metodo per preparare il testo per l'input per un processo di machine-learning.

BoW elenca le parole associate ai conteggi delle parole per documento. Nella tabella in cui sono archiviate le parole e i documenti che diventano effettivamente vettori, ogni riga è una parola, ogni colonna è un documento e ogni cella è un conteggio delle parole.

Ciascuno dei documenti nel corpus è rappresentato da colonne di uguale lunghezza. Questi sono vettori di conteggio di parole (2).

BOW è un approccio ampiamente utilizzato con:

- Elaborazione del linguaggio naturale;
- Recupero di informazioni da documenti;
- Classificazioni dei documenti.

Ad un livello elevato, comporta i seguenti passaggi:

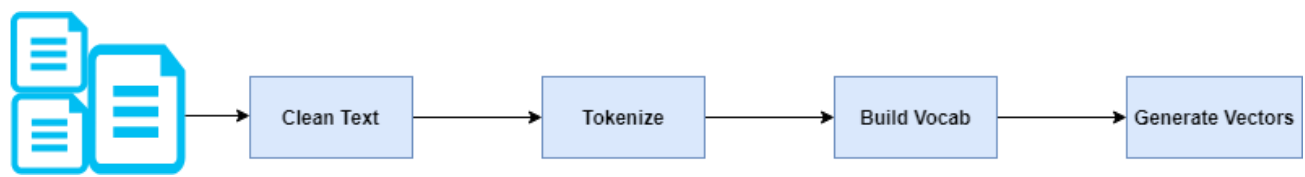

*Figura 3.3 - Pipeline delle BOW*

### 2.5.1 Limitazioni di BoW

- **Significato semantico**: l'approccio BoW di base non considera il significato della parola nel documento. Ignora completamente il contesto in cui viene utilizzato. La stessa parola può essere utilizzata in più punti in base al contesto o alle parole vicine.
- **Dimensione vettoriale:** per un documento di grandi dimensioni, la dimensione del vettore può essere enorme con conseguente calcolo e tempo. Potrebbe essere necessario ignorare le parole in base alla pertinenza relativamente al caso d'uso.

### 2.5.2 Tokenize

Il primo passo procedurale sarà quello di ridurre il numero di dimensioni dei vettori, questo si traduce nella rimozione delle stop-words.

- Le **stop-word** sono parole che non contengono informazioni utili, al contrario aumenterebbero l'entropia nella frase.
- La **tokenizzazione** è l'atto di spezzare una sequenza di stringhe in pezzi come parole, parole chiave, frasi, simboli e altri elementi chiamati token. I token possono essere singole parole, frasi o persino frasi intere. Nel processo di tokenizzazione, alcuni caratteri come i segni di punteggiatura vengono scartati.
- La **lemmatizzazione** prende in considerazione l'analisi morfologica delle parole. Per fare ciò, è necessario disporre di dizionari dettagliati che l'algoritmo può esaminare per collegare la forma flessa al suo lemma.

## 2.6 Calcolo della distanza del coseno

I vettori delle frasi che formano la knowledge-base costituiscono la matrice K, applichiamo lo stesso procedimento di vettorializzazione sulla frase F che intendiamo classificare e per la quale vogliamo misurare l'appartenenza al dominio. Quindi da una parte abbiamo il vettore che rappresenta F e dall'altro la matrice K.

Nel caso in cui il dizionario non contenesse delle parole sarà necessario, prima di poter effettuare il confronto di F con la matrice K, arricchire K aumentando le dimensioni dei vettori codificati.

Per far questo si andrà ad aggiungere tante dimensioni quante sono le parole che non sono presenti nel dizionario, popolando la rispettiva frequenza moltiplicata per un fattore di penalità pari a 2.5 (valore scelto a seguito di una fase di test), e aumentando la dimensione dei vettori della knowledge-base, impostando l'occorrenza pari a 0.

Ad esempio nel caso in cui si volesse calcolare la similarità di un documento rispetto a una knowledge-base nel dominio Telco, si calcolerà il vettore della frase, utilizzando il "Modello Bag of words" ottenendo il vettore come di seguito.

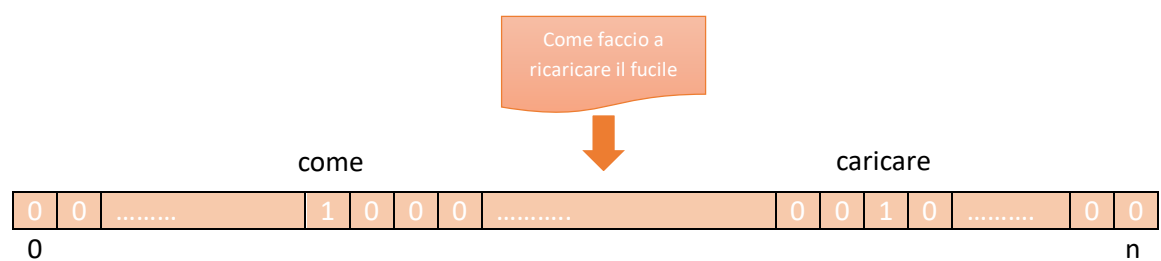

Dove ogni dimensione, ovvero ogni elemento del vettore, equivale a una parola del dizionario e al suo interno è presente il numero di volete che la parola è presente nella frase.
Successivamente si andrà ad aggiungere un ulteriore dimensione che identificherà la parola non presente nel dizionario, e si moltiplicherà la sua occorrenza per un valore di penalizzazione pari a 2.5 come di seguito.

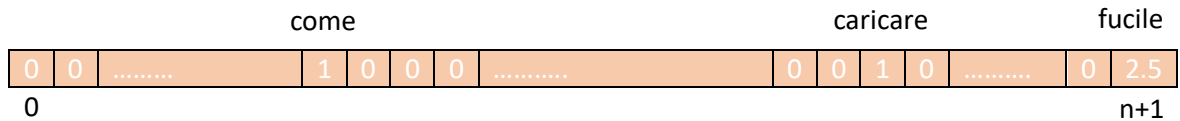

Mentre la matrice della knowledge base contenete le frasi da confrontare incrementerà il numero di colonne, così da uniformare la dimensione delle colonne tra il vettore e la matrice, come indicato di seguito.

$$\begin{bmatrix} a_{11} & \cdots & a_{1n} \\ \vdots & \ddots & \vdots \\ a_{n1} & \cdots & a_{nn} \end{bmatrix} \rightarrow \begin{bmatrix} a_{11} & \cdots & a_{1n+i} \\ \vdots & \ddots & \vdots \\ a_{n1} & \cdots & a_{nn+i} \end{bmatrix}$$

Con $i = 1$ dove 1 è il numero di parole non presenti nella knowledge base nel esempio.

Questa manipolazione è necessaria per poter incrementare la distanza dei vettori messi a confronto, nel caso in cui ci fossero delle parole non presenti nel dizionario di dominio. Infatti una frase che contiene una o più parole non presenti nel dizionario ha maggiori possibilità di non appartenere al dominio.
In una sequenza di tipo iterativo applicata alla matrice K, confronteremo, utilizzando il calcolo della distanza del coseno (formula presentata nel capitolo 2.3), il vettore di F l'n-esimo vettore in K.

Il vettore F apparterrà al dominio solo se presa la più piccola distanza del coseno sarà minore o uguale di 0.5, valore scelto empiricamente.

## 2.7 Implementazione e librerie

La classificazione del dominio della frase tramite il calcolo del Cosine Similarity è stata implementata su un cluster Apache Ambari[2] e tramite l'utilizzo del framework Apache Spark, questa scelta è dovuta alla potenza di calcolo necessaria per poter effettuare il confronto di un determinato vettore con un knowledge-base contenente i vettori delle frasi di un determinato dominio.

Questa operazione in un ambiente mono-macchina potrebbe impiegare un tempo considerevole. Sfruttando il cluster di Ambari e la libreria di Spark si riescono a paragonare 3000 vettori in poco meno di 2 secondi.

La libreria utilizzata per poter effettuare il calcolo della distanza del coseno è ***mahout-core,*** ipotizzando l'uso di Apache Maven come repository di librerie, per poter utilizzare i metodi sarà necessario inserire nel pom.xml la seguente dipendenza:

```xml
<dependency>
        <groupId>org.apache.mahout</groupId>
        <artifactId>mahout-core</artifactId>
        <version>0.9</version>
</dependency>
```

Di seguito, invece, viene descritto il metodo per il calcolo della similarità implementato in java, gli argomenti del metodo sono la frase da classificare, la knowledge-base, lo SparkSession, il JavaSparkContext e la lista delle categorie.

La logica alla base del metodo è incentrata nello scorrere il dataset contenete i vettori tramite espressione lambda, così da poter paragonare i vari vettori con il vettore da classificare.

Si evince che nella lambda viene utilizzato il metodo ".distance" della classe CosineDistanceMeasure che restituisce la distanza del coseno, e restituirà un dataset contenente la distanza calcolata con il relativo vettore. Ogni vettore contiene un determinato "label" che rappresenta la categoria a cui appartiene.

---

[2] Il cluster utilizzato è composta da 14 macchine omogenee così descritte: VMWARE virtual machines(MB Intel 440BX, Intel Xeon Gold 6140@2.30GHz DualCore, 16GB RAM, 140GB 53c1030 PcI-X Fusion-MPT HDD)

```java
public static JsonObject getKNN(double[] phrase,
                                Dataset<Row> inputDF,
                                SparkSession spark,
                                JavaSparkContext jsc,
                                List<String> classList) {

    double[] vectPhrase = Vectors.dense(phrase).toArray();
    Dataset<Row> minCosDataset = inputDF.map(row -> {
                                    double[] features = ((Vector) row.getAs("features")).toArray();
                                    double dist = CosineDistanceMeasure.distance(vectPhrase, features);
                                    return RowFactory.create(row.getAs("label"), dist);
                                }, KNNSchema).sort("CosDist");

    double minCosineDistance = 1 -((Double) minCosDataset.first().getAs("CosDist"));
    int label = ((Double) minCosDataset.first().getAs("label")).intValue();
    List<Double> knnResult = new ArrayList<Double>(Collections.nCopies(classList.size(), 0.0));
    knnResult.set(label, 1.0);

    JsonObject jsKnn = new JsonObject();
    jsKnn.addProperty("similarityValue", minCosineDistance);
    jsKnn.add("knnResult",Utils.getJsonUtility().fromObjectToJson(knnResult));
    return jsKnn;
}
```

## 3. Conclusioni

In conclusione la scelta effettuata per verificare il dominio della frase ha portato a dei risultati molto soddisfacenti, partendo da una base documentale non molto grande, ovvero 300 frasi per categoria, per un totale di 10 categorie.

La scelta di un metodo di classificazione che non prevede una creazione del modello permette la classificazione a runtime della sua appartenenza. Lo svincolarsi dalla creazione del modello permette di incrementare le dimensioni del vettore così da poter aumentare la precisione di predizione di appartenenza al dominio.

Questa scelta però porta con sé degli svantaggi come ad esempio l'aumento delle risorse utilizzate per il calcolo e di conseguenza l'aumento dell'elaborazione con l'aumentare delle quantità dei vettori contenuti nella knowledge-base.

Di contro i vantaggi che avvalorano la scelta di questo modello non sono da poco, in particolare rende semplice sia l'implementazione del suo algoritmo che l'inserimento o l'espansione del corpus documentale.

Il confronto fra i pro e i contro dell'utilizzo di questo algoritmo e i risultati ottenuti hanno portato a scegliere il suddetto algoritmo per il calcolo della similarità.

Infatti i test effettuati hanno riscontrato una accuracy pari al 98,40%, utilizzando la distanza del coseno, questo risultato è il migliore rispetto all'utilizzo di altri metodi come si evince nel Tabella 2.1.

# 4. Bibliografia